\documentclass{ifacconf}
\usepackage{xcolor}
\usepackage{graphicx} 

\usepackage{natbib}        
\usepackage{booktabs}
\usepackage{multirow}
\usepackage{multicol}
\usepackage{amsmath}
\usepackage{dblfloatfix}
\usepackage{amssymb}
\usepackage{mathtools}
\usepackage{flushend}

\usepackage{color}

\makeatletter
\let\old@ssect\@ssect 
\makeatother
\usepackage{hyperref}
\makeatletter
\def\@ssect#1#2#3#4#5#6{%
  \NR@gettitle{#6}
  \old@ssect{#1}{#2}{#3}{#4}{#5}{#6}
}
\makeatother
\begin{document}
\mathtoolsset{showonlyrefs}
\begin{frontmatter}

\title{Robustness Benchmark of Road User Trajectory Prediction Models for Automated Driving\thanksref{footnoteinfo}} 


\thanks[footnoteinfo]{This work was supported by SAFE-UP under EU's Horizon 2020 research and innovation programme, grant agreement 861570.}

\author[First]{Manuel Muñoz Sánchez} 
\author[First,Second]{Emilia Silvas}
\author[First,Third]{Jos Elfring} 
\author[First]{René van de Molengraft}

\address[First]{Department of Mechanical Engineering, TU/e, Eindhoven, NL.}
\address[Second]{Department of Integrated Vehicle Safety, TNO, Helmond, NL.}
\address[Third]{VDL CropTeq Robotics, Eindhoven, NL.}

\begin{abstract}                
Accurate and robust trajectory predictions of road users are needed to enable safe automated driving.  To do this, machine learning models are often used, which can show erratic behavior when presented with previously unseen inputs. 
In this work, two environment-aware models (MotionCNN and MultiPath++) and two common baselines (Constant Velocity and an LSTM) are benchmarked for robustness against various perturbations that simulate functional insufficiencies observed during model deployment in a vehicle: unavailability of road information, late detections, and noise. 
Results show significant performance degradation under the presence of these perturbations, with errors increasing up to +1444.8\% in commonly used trajectory prediction evaluation metrics. Training the models with similar perturbations effectively reduces performance degradation, with error increases of up to +87.5\%.  
We argue that despite being an effective mitigation strategy, data augmentation through perturbations during training does not guarantee robustness towards unforeseen perturbations, since identification of all possible on-road complications is unfeasible. Furthermore, degrading the inputs sometimes leads to more accurate predictions, suggesting that the models are unable to learn the true relationships between the different elements in the data.
\end{abstract}


\end{frontmatter}

\section{Introduction}
Trajectory prediction of road users (RUs) for automated vehicles (AVs) has received much attention in recent years, since it allows an AV to anticipate how the environment will evolve and react better, achieving safer, more comfortable, and more efficient reactions. Given the success of machine learning in various fields, a data-driven approach has also been adopted in trajectory prediction for AVs. 

The main focus of early works on trajectory prediction was to increase predictive accuracy leveraging as much information as possible about the environment, such as surrounding RUs \citep{Alahi2016SocialSpaces,Kosaraju2019Social-BiGAT:Networks} and road infrastructure \citep{Salzmann2020Trajectron++:Data,Yoon2020Road-awareHighways}. To develop these methods, a heavily processed dataset is often used, where detailed information about the RUs and road geometry is available. To deal with more realistic data and previously unseen situations, recently more emphasis is being placed on robustness of these methods \citep{Roelofs2022CausalAgents:Relationships,Cao2022AdvDO:Prediction,Bahari2022VehicleEverywhere,Saadatnejad2022AreSocially-aware,Zhang2022OnVehicles}, since they must achieve reliable predictions.

An obvious approach to evaluate and increase model robustness towards unseen situations is to collect more data containing those missing situations (e.g. different locations, weather conditions, faulty sensors, etc.), which is costly, time consuming, and not scalable. To address this lack of data, a common practice is to introduce \textit{perturbations} (i.e. synthetic variations) to an already existing dataset to verify that the model still produces sensible predictions. 
Perturbations can be either \textit{complementing}, which aim to enhance the recorded data with previously unseen situations (e.g.craft fake historical trajectories \citep{Cao2022AdvDO:Prediction,Zhang2022OnVehicles}), or \textit{disruptive}, which simulate complications such as sensor noise \citep{Zamboni2022PedestrianNetworks}. 

One of the limitations of current robustness evaluations through disruptive perturbations is that the choice of perturbations does not resemble harsh realistic conditions (e.g. longer data losses or high noise levels). These can occur and must be considered if AVs should function in various challenging environments (e.g. adverse weather, poor lighting, or infrastructural changes). When simulating disruptive perturbations, it is common to assume a \textit{bias} in the observed trajectories, or sample artificial observation noise from some distribution, often Gaussian \citep{Zamboni2022PedestrianNetworks}. However, the motivation behind the choice of parameters for sampling this noise is often lacking. If the analyzed perturbations are not representative of what one might encounter in reality, the identified performance degradation will not be representative of what one can expect in the vehicle. Consequently, the mitigation approaches will not be effective in reality. 

To bridge the gap between heavily processed rich learning datasets and in-vehicle data, in this work, we perform a robustness evaluation of several trajectory prediction models introducing extreme possible perturbations.  
Such perturbations are complete unavailability of road information, late detections with only one observation, and highly noisy heading angle measurements, which could be caused by faulty sensors or adverse weather conditions~\citep{Zang2019TheCar}.
To show the impact these perturbations can have in data-driven models that did not consider similar perturbations during training, we first assess the performance degradation using only perturbed data. Next, we train the models considering these perturbations and show the robustness increase towards perturbed data and potential performance decrease when using original data. 

Thus, our work's main contribution is a benchmark that assesses the robustness of various machine learning trajectory prediction models against severe and highly detrimental perturbations. The choice of specific perturbations is motivated by real functional insufficiencies in world modeling that we observed when deploying our models in a vehicle. Additionally, we encourage the scientific community to perform similar types of robustness studies by sharing the framework we created for our study as reference\footnote{\url{https://bit.ly/IFAC23-robustness-benchmark}}.


The remainder of this article is structured as follows. Section~\ref{sec:trajectory-prediction} presents the trajectory prediction problem and summarizes related work on robustness of trajectory prediction models. Section~\ref{sec:method} outlines the benchmark procedure and introduces the perturbations considered. Section~\ref{sec:results} summarizes the results, and Section~\ref{sec:conclusion} concludes the work and highlights future improvements.


\section{Preliminaries} \label{sec:trajectory-prediction}
\subsection{Trajectory Prediction}
Trajectory prediction refers to predicting the future positions of an RU. Given a trajectory prediction model $\mathcal{M}$ and its inputs $\mathcal{I}$, it will generate future predictions $\mathcal{P} = \mathcal{M}(\mathcal{I})$. The types of inputs and predictions are dependent on the model, although to generate predictions for a target RU $n \in N$, at least $\mathcal{I} = (\mathbf{s}^n_{t_0}, T, *)$ and $\mathcal{P} = (\mathbf{\hat{x}}^n_T, *)$, where $\mathbf{s}^n_{t_0}$ denotes the current state of $n$ at time $t=0$, $T$ denotes the desired prediction horizons, and $\mathbf{\hat{x}}^n_T$ denotes one or more predictions of future positions of n at those horizons. Additionally, $*$ denotes other optional inputs and outputs. 

Typically, additional inputs are a map $\mathbf{m}$ containing geometric and semantic information of the static environment (e.g. a road model); past states of the target $\mathbf{s}^n_{t::0}$ from time $t<0$ up to but not including $t=0$; and other RUs' states, $\mathbf{s}^{N \setminus n}_{t:0}$ from time  $t<0$ up to and including $t=0$. Typical additional outputs are some measure of uncertainty $\mathcal{U}$ associated with the predictions. 

To assess the accuracy of a trajectory prediction model, its predictions at times $T$ for a set of RUs $N$, $\mathcal{P} = (\mathbf{\hat{x}}^N_T, *)$, are compared with the real (ground truth) future positions, $\mathbf{x}^N_T$. To quantify this accuracy, one of the most common metrics is minADE~\citep{Rasouli2020DeepSurvey},
defined as  
\begin{equation}\label{eq:minADE}
    \text{minADE}(\mathbf{\hat{x}}^N_T, \mathbf{x}^N_T) =  \sum_{n\in N} \min_{\mathbf{y} \in \mathbf{\hat{x}}^n_{T}} \sum_{\substack{t \in T }} \frac{\Vert \mathbf{y}_t - \mathbf{x}_t^n \Vert_2}{|N| \cdot |T|},
\end{equation}
where $|.|$ denotes the size of a set, and $\Vert . \Vert_2$ denotes the L2-norm of a vector. Note that this metric only considers $\hat{\mathbf{x}}$ and not $\mathcal{U}$. Despite this limitation, it has become widely popular since it allows direct comparison between models that provide $\mathcal{U}$ and those that do not.


\subsection{Trajectory Prediction Robustness}
To enable proactive automated driving behaviour that is safe and comfortable, predicting the behavior of RUs is crucial, even when facing challenging driving conditions or faulty data. To evaluate and increase robustness of prediction models, \textit{perturbations} are often used \citep{Roelofs2022CausalAgents:Relationships}.




\textit{Perturbations}{ introduce synthetic variations to testing data, $\mathcal{I'} \sim \mathcal{I}_{test}$, to simulate unrecorded data. 
Perturbations can be \textit{complementing}, or \textit{disruptive}. Complementing perturbations aim to simulate unrecorded data which can be expected in reality. For instance \cite{Roelofs2022CausalAgents:Relationships} show how the removal of irrelevant agents from the scene (e.g. parked vehicles) has a detrimental effect on predictive accuracy, and \cite{Zhang2022OnVehicles} show that subjecting the models to adversarial attacks (i.e. crafting new realistic historical trajectories) leads to lower accuracy. Disruptive perturbations aim to lower the reliability of the input data, most typically by simulating noise \citep{Zamboni2022PedestrianNetworks}. 
Increasing robustness towards perturbations is often addressed introducing similar perturbations to the training data, $\mathcal{I}_{\text{train}}$ \citep{Zhang2022OnVehicles,Zamboni2022PedestrianNetworks}.}

Our work falls under the category of disruptive perturbations. In particular, we evaluate robustness towards extreme perturbations that we observed while deploying prediction models in a real vehicle (i.e. complete road unavailability, late detections with only one observation, and highly noisy heading measurements).
Additionally, we investigate the effectivity of introducing similar perturbations during training.

\subsection{Data Preparation for a Trajectory Prediction Model} \label{sec:data-prep}
\begin{figure}[b]
    \centering
    \includegraphics[width=\linewidth]{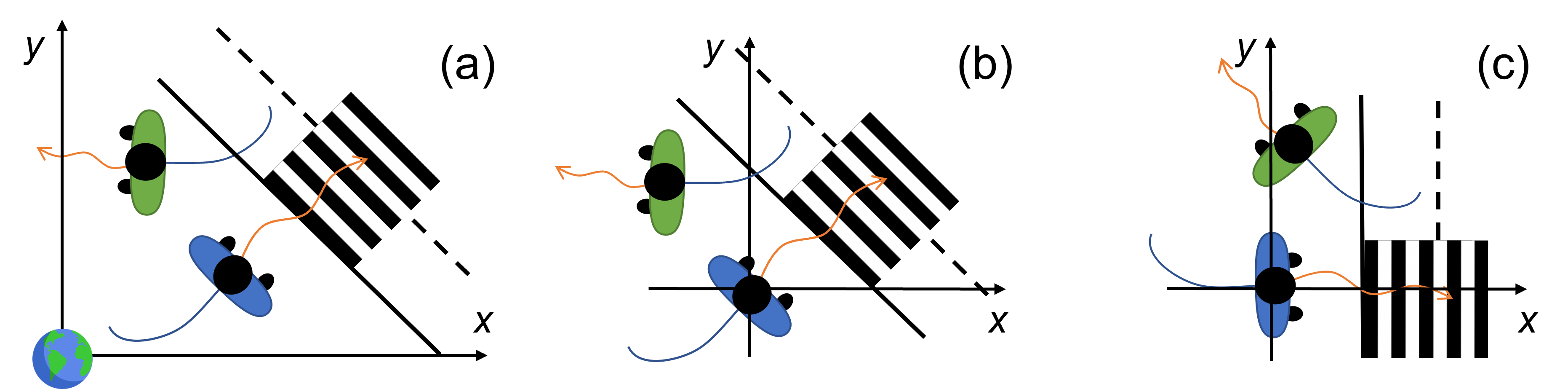}
    \caption{Data preprocessing of trajectory prediction models.}
    \label{fig:preprocess}
\end{figure}

%
To reduce model complexity and required training time during model development, it is common practice to modify its inputs $\mathcal{I}$ before prediction 
with a 2-step process:

Firstly, instead of operating on world or AV coordinates (Fig.~\ref{fig:preprocess}a), the RU's reference frame is used, shifting the scene to be centered on the last observed target position (Fig~ \ref{fig:preprocess}b), and then rotating it by the current RU's heading angle (Fig~ \ref{fig:preprocess}c). Secondly, all features are scaled to a common range, or to achieve zero mean and unit variance.

\section{Methodology \& Experiments}~\label{sec:method}
\vspace{-25pt}
\subsection{Benchmarked Models}
Four models are evaluated: constant velocity (CV); an LSTM encoder-decoder similar to \cite{MunozSanchez2022}, MotionCNN \citep{Konev2021MotionCNN:Driving}, and MultiPath++ \citep{Varadarajan2022MultiPath++:Prediction,Konev2022MPA:Prediction}. CV and LSTM are common baselines, and MotionCNN and MultiPath++ are state of the art models leveraging road geometry and surrounding RUs in their predictions. 

\subsubsection{Constant Velocity}
To make predictions for RU $n$ with CV, the only required input is its current state consisting of positions and velocities, $\textbf{s}_t = (\mathbf{x}_t,\mathbf{v}_t)$ for $t=0$ (i.e. current time). Thus $\mathcal{I} = (\mathbf{s}^n_0)$. Then, for a given prediction horizon $t'$, its future positions are computed as
\begin{equation}
    \hat{\textbf{x}}^n_{t+t'} = \textbf{x}^n_t + t' \cdot \textbf{v}^n_t.
\end{equation}

\subsubsection{LSTM}
A vanilla LSTM encoder-decoder with 3 layers of 128 neurons for the encoder and decoder that does not consider road information or surrounding RUs.  With LSTM, the state of an RU at time $t$ is given by
\begin{equation}
\mathbf{s}_t = (\mathbf{x}_t, \theta_t, \mathbf{v}_t, w, l, \nu_t, \boldsymbol{\tau}),
\end{equation}
where $\theta$ denotes heading angle; $w$ and $l$ width and length; $\nu \in \{0, 1\}$ indicates validity (e.g. the target was temporarily occluded); and $\boldsymbol{\tau}$ is a one-hot encoding indicating the RU type (i.e. unset, vehicle, pedestrian, cyclist, or other). 
In the data we used, the length of historical and future observations are 1 and 8 seconds respectively, sampled at 10Hz. Thus, to make predictions of RU $n$ at time $t$ with LSTM, its inputs are $\mathcal{I} = (\mathbf{s}^n_{t-1:t}, T)$, where 
\begin{equation}\label{eq:T}
    T = \{\frac{t}{10} \mid 0 < t \leq 80 \land t \in \mathbb{Z}\}. 
\end{equation}
Its predictions $\mathcal{P} = (\mathbf{\hat{x}}^n_{T})$ are produced recursively given the previous position and the network's hidden and cell states~\citep{Park2018Sequence-to-SequenceArchitecture}.

\subsubsection{MotionCNN}
A convolutional neural network-based architecture~\citep{Konev2021MotionCNN:Driving}. Its inputs are raster images of 224x224 pixels with 25 channels, where the first three channels encode road geometry, the next eleven encode the past and current positions of the target, and the last 11 all other RUs. Thus, to make predictions of RU $n \in N$ at time $t$, MotionCNN uses $\mathcal{I} = (\mathbf{s}^N_{t-1:t}, T, \mathbf{m})$ with $T$ as in \eqref{eq:T}, and makes predictions $\mathcal{P} = (\mathbf{\hat{x}}^n_T,\mathbf{p})$ with probabilities $\mathbf{p}$ of the predicted trajectories.

\subsubsection{MultiPath++}
The state of an RU at time $t$ is given by 
\begin{equation}
\mathbf{s}_t = (\mathbf{x}_t, \theta_t, u_t, w, l, \nu_t, \tau, \Delta\mathbf{x}_t, \Delta\theta_t, \Delta u_t, \Delta \nu_t),
\end{equation}
where $u$ denotes speed, and $\Delta$ denotes the change in the variable that follows with respect to the previous time. Thus, to make predictions of RU $n \in N$ at time $t$, MultiPath++ uses $\mathcal{I} = (\mathbf{s}^N_{t-1:t}, T, \mathbf{m})$ with $T$ as in \eqref{eq:T}, and makes predictions $\mathcal{P} = (\mathbf{\hat{x}}^n_T,\mathbf{p},\Sigma)$ with probabilities $\mathbf{p}$ and covariance matrix $\Sigma$ of the predicted trajectories.

\subsection{Baseline Evaluation}\label{sec:base-eval}
To establish a baseline performance, we evaluate the models under nominal conditions (i.e. training and evaluating them with the original data). To that end, we use the Waymo Open Motion Dataset (WOMD) \citep{Ettinger2021LargeDataset}. To reduce computational load, we only used trajectories of RUs labeled \textit{tracks to predict}, which feature more diverse behavior than the rest~\citep{Ettinger2021LargeDataset}.
Since the ground truth for the test data is kept hidden for motion prediction challenges\footnote{\url{https://waymo.com/open/challenges/}}, we reserved 1/3 of the original validation split for testing. Thus, our testing data consists of trajectories of 54903 vehicles, 6958 pedestrians and 1784 cyclists.
Prediction accuracy is reported by means of the predictions' minADE, as defined in \eqref{eq:minADE}, over all prediction horizons $T$ as defined in \eqref{eq:T}. 
\begin{figure*}[ht]
    \centering
    \includegraphics[width=1\linewidth]{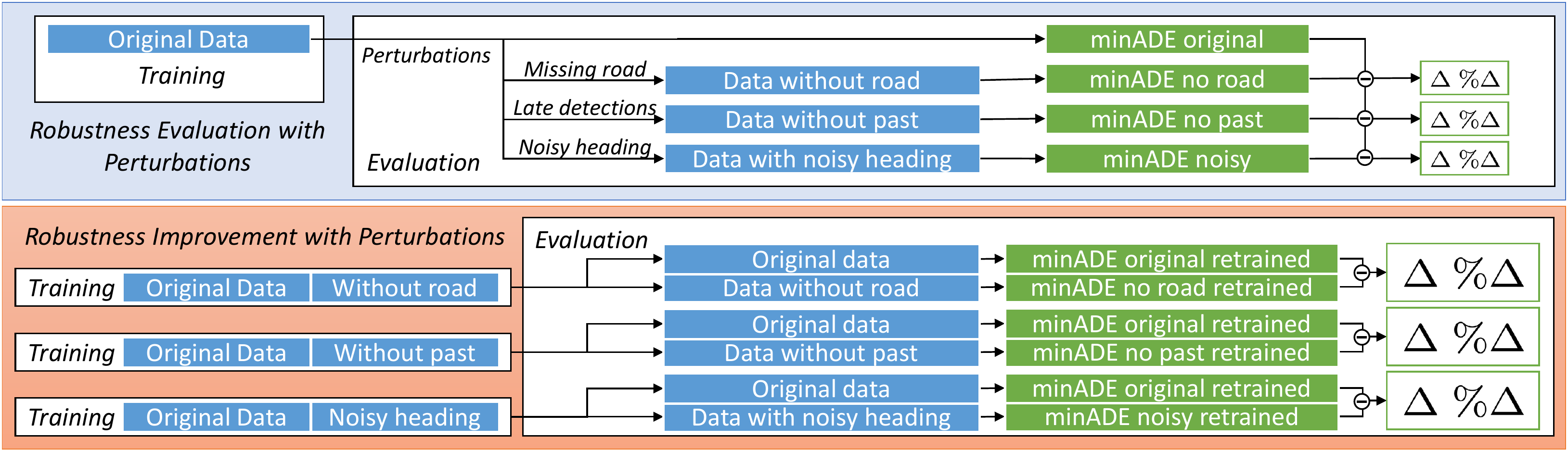}
    \vspace{-15pt}
    \caption{Overview of the benchmarking procedure to assess the robustness of models towards different perturbations.}
    \label{fig:overview}
\end{figure*}
\begin{table*}[b]
\begin{center}
\caption{Model minADE with original data and perturbed variants}\label{tab:evaluation-perturbances}
\begin{tabular}{@{}cc|ccc|ccc|ccc@{}}
\toprule
\multicolumn{1}{c|}{Model} & Original & \multicolumn{3}{c|}{Missing Road} & \multicolumn{3}{c|}{Late Detection} & \multicolumn{3}{c}{Noisy Heading} \\
\multicolumn{1}{l|}{}            &      &      & $\Delta$ & \%$\Delta$ &      & $\Delta$ & \%$\Delta$ &       & $\Delta$ & \%$\Delta$ \\ \hline
\multicolumn{1}{c|}{CV}          & 8.83 & -    & -        & -          & -    & -        & -          & -     & -        & -          \\
\multicolumn{1}{c|}{LSTM}        & 5.16 & -    & -        & -          & 9.44 & 4.28     & 82.95\%    & 23.45 & 18.29    & 354.46\%   \\
\multicolumn{1}{c|}{MotionCNN}   & 1.39 & 2.93 & 1.54     & 110.79\%   & 3.96 & 2.57     & 184.89\%   & 18.74 & 17.35    & 1248.20\%  \\
\multicolumn{1}{c|}{MultiPath++} & 1.34 & 2.3  & 0.96     & 71.64\%    & 2.49 & 1.15     & 85.82\%    & 20.7  & 19.36    & 1444.78\%  \\ \bottomrule
\end{tabular}
\end{center}
\end{table*}

\setlength{\tabcolsep}{3pt}
\begin{table*}[htb]
\begin{center}
\centering \scriptsize
\caption{Model minADE after training with original and perturbed data}\label{tab:evaluation-retrained} 
\begin{tabular}{@{}ccccc|cccc|cccc@{}}
\toprule
\multicolumn{1}{c|}{Model} & \multicolumn{4}{c|}{Missing Road} & \multicolumn{4}{c|}{Late Detection}            & \multicolumn{4}{c}{Noisy Heading}     \\
\multicolumn{1}{c|}{} &
  Original &
  Perturbed &
  $\Delta$ &
  \%$\Delta$ &
  Original &
  Perturbed &
  $\Delta$ &
  \%$\Delta$ &
  Original &
  Perturbed &
  $\Delta$ &
  \%$\Delta$ \\ \hline
\multicolumn{1}{c|}{LSTM}  & -     & -     & -    & -    & 5.17 (+0.19\%) & 9.37 & 4.20 & 81.24\% & 5.39 (+4.46\%) & 6.3 & 0.91 & 16.88\% \\
\multicolumn{1}{c|}{MotionCNN} &
  1.35 (-2.88\%) &
  1.59 &
  0.24 &
  17.78\% &
  1.34 (-3.60\%) &
  1.75 &
  0.41 &
  30.60\% &
  1.45 (+4.32\%) &
  1.46 &
  0.01 &
  0.69\% \\
\multicolumn{1}{c|}{MultiPath++} &
  1.56 (+16.42\%) &
  1.81 &
  0.25 &
  16.03\% &
  1.28 (-4.48\%) &
  2.4 &
  1.12 &
  87.50\% &
  1.33 (-0.75\%) &
  1.31 &
  -0.02 &
  -1.50\% \\ \bottomrule
\end{tabular}
\end{center}
\end{table*}

\subsection{Robustness Evaluation with Perturbations}
Based on our experience deploying trajectory prediction models in a vehicle, we focus on three 
highly detrimental perturbations that can render the models unusable.
Such perturbations are unavailability of road information; late detections; and highly noisy RU's heading angle.

\subsubsection{Missing road information}
Availability of road information is not guaranteed. If models are always trained with it, and it is temporarily missing, the results could be unpredictable. To evaluate this case, all road information is removed.


\subsubsection{Late detections}
When perturbing the dataset to simulate perceptual loss, it is common to drop some observations of the historical trajectories with certain probability \citep{Konev2022MPA:Prediction}. This way, even if there are gaps in the observed trajectories, the models can exploit the remaining historical data to produce better predictions. With late detections, however, there is no historical data to exploit, thus prediction accuracy will naturally suffer. To investigate the potential performance degradation when lacking past observations, only the most recent observation is considered.

\subsubsection{(Highly) noisy heading} 
It is common practice to transform the inputs to a trajectory prediction model relative to the RU's coordinate frame, as introduced in Section~\ref{sec:data-prep}. If the last observed RU state is highly noisy, the entire scene will be transformed erroneously, which can have a highly detrimental effect on the predictions. 
During model deployment in the vehicle, higher heading angle errors than anticipated were observed, resulting in invalid predictions because the models were never trained with similar noise levels. To evaluate robustness towards extreme cases, we introduce a 90 degree offset in the last observed target heading angle.

As illustrated in Fig.~\ref{fig:overview}(top), for each perturbation introduced we report its minADE averaged over all trajectories using the perturbed dataset, its performance degradation $\Delta$, and its relative increase $\%\Delta$ given by 
\begin{equation}
    \Delta = \text{minADE\_perturbed} - \text{minADE\_original},
\end{equation}
\begin{equation}
    \%\Delta = \frac{\Delta}{\text{minADE\_original}}.
\end{equation}
Additionally, to investigate the impact of these perturbations for each prediction, we also analyze the change of minADE per trajectory.

\subsection{Robustness Improvement with Perturbations} 
For each perturbation, a new dataset is generated consisting of the original data and a perturbed copy. As shown in Fig.~\ref{fig:overview}(bottom), a new version of the models is then trained with each of the new perturbed datasets, after which a similar evaluation is performed to assess the robustness improvement. Additionally, we report the minADE relative increase of each model evaluated on the original data after it is trained with perturbed data.

\section{Results}\label{sec:results}
\subsection{Robustness Towards Perturbations}
Table \ref{tab:evaluation-perturbances} shows the performance of each model when provided with the original and perturbed data. As expected, MotionCNN and MultiPath++ clearly outperform CV and LSTM, since they leverage environmental information such as road geometry and surrounding RUs. Additionally, Fig.~\ref{fig:per-example}(top) shows a density plot of $\Delta$ for each individual prediction instead of averaged over all trajectories. Finally, Fig.~\ref{fig:example-predictions}(top) shows examples of predictions produced by MultiPath++ given the original and perturbed data.

\subsubsection{Missing road information} CV and LSTM do not use road information, and are therefore unaffected by its absence. On the other hand, MotionCNN and MultiPath++ suffer a severe performance drop (+110.79\% and +71.64\% respectively), although they still outperform the environment-unaware models. Despite this average performance drop, there are several cases where prediction accuracy improves after removing the road, as seen by the negative $\Delta$ in Fig.~\ref{fig:per-example}a.
Fig.~\ref{fig:example-predictions}b shows an example of a prediction where the lack of road information has a negative impact, causing off-road predictions. 

\subsubsection{Late detections} 
\begin{figure*}[tb]
    \centering
    \includegraphics[width=\linewidth]{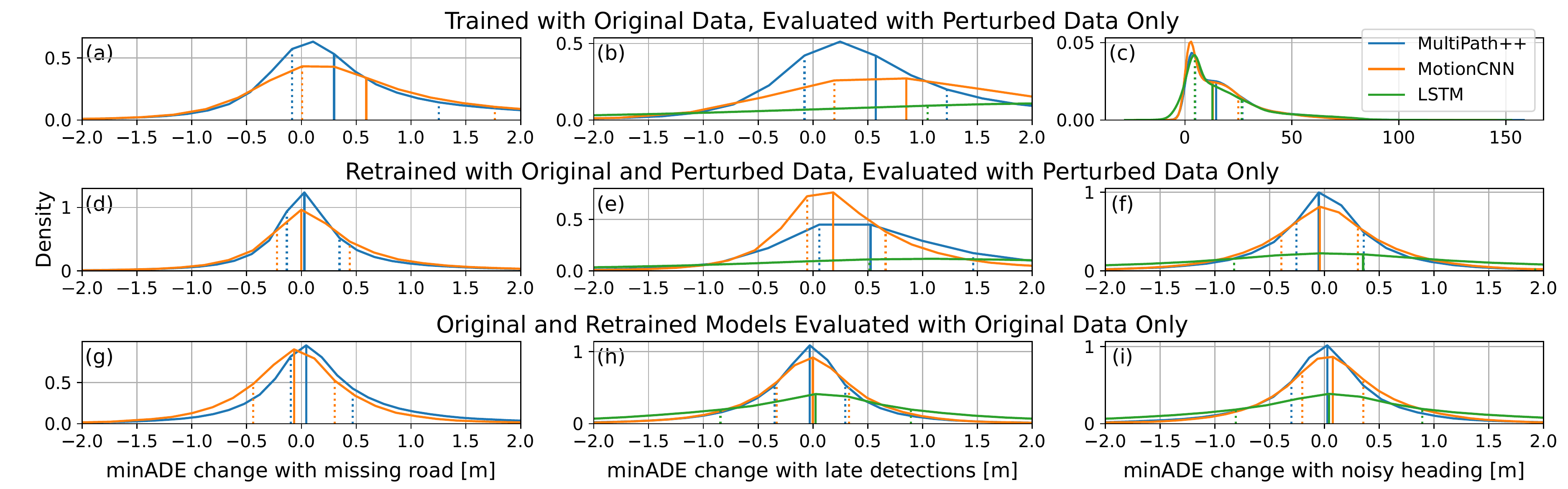}
    \vspace{-12pt}
    \caption{Density plots of minADE change per trajectory. Top: $\Delta$ of original models evaluated on perturbed data. Middle: $\Delta$ of retrained models evaluated on perturbed data. Bottom: minADE difference of original and retrained models when evaluated on original data. Solid vertical lines denote the median, and dotted lines the 25th and 75th percentiles. Note: horizontal axis is fixed to [-2,2]m in most plots, and the long tails could extend further than shown.
    }
    \label{fig:per-example}
\end{figure*}

\begin{figure*}[b]
    \centering
    \includegraphics[width=0.99\linewidth]{{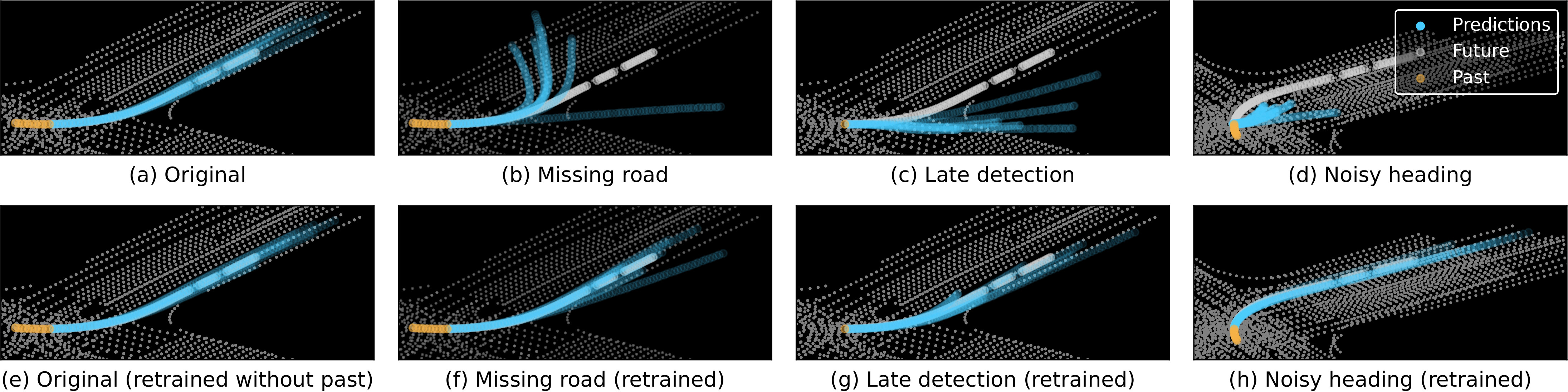}}
    \vspace{-6pt}
    \caption{Examples of MultiPath++ predictions on original and perturbed data (scene 106fb050cdf836af, agent 1117). Trained and tested on original data (a), trained on original data and tested with pertubed data (b-d), trained with both original and perturbed data (e-h), and tested with original (e) and perturbed data (f-h).}
    \label{fig:example-predictions}
\end{figure*}

CV only uses the most recent observation, and as such is unaffected by lack of past observations. LSTM and MultiPath++ undergo an error increase of approximately +85\%, resulting in 
higher errors than those of CV in the case of LSTM. Since MotionCNN's raster inputs do not contain information about the target's current velocities, it is unable to infer it from its past positions and its minADE almost triples (+184.89\%). Despite having road information, the models sometimes give predictions going off-road (Fig.~\ref{fig:example-predictions}c) due to lack of historical data, which suggests the model was unable to infer behavior that is obvious to humans: vehicles drive on the road (at least in nominal conditions). Nevertheless, under this perturbation there are also a significant number of predictions for which minADE improves (Fig.~\ref{fig:per-example}b).

\subsubsection{Noisy heading angle} CV does not use heading angle, as the velocities in each direction are provided as input directly, therefore it remains unaffected by this perturbation. All machine learning models suffer an extreme performance degradation of up to +1444.78\%, resulting in significantly worse performance than that of CV. This performance degradation is to be expected, since the models have been trained modifying their inputs as described in Section \ref{sec:base-eval} to reduce model complexity and training times, and if the entire scene is rotated with a wrong angle, they are unable to produce sensible predictions. 
%

\subsection{Robustness Towards Perturbations after Retraining}
Table \ref{tab:evaluation-retrained} shows the performance of each machine learning model when provided with the original data and the three perturbations after being trained with original and perturbed data. Additionally, when evaluated using the original dataset, the performance increase with respect to the model that was trained without perturbations is reported. Fig.~\ref{fig:per-example}(middle) summarizes $\Delta$ of each individual prediction. Finally, Fig.~\ref{fig:example-predictions}(bottom) shows example predictions with a retrained model.

\subsubsection{Missing road information}
Applying this perturbation during training leads to different results. MotionCNN not only achieves lower performance degradation on perturbed data (+17.78\% vs. +110.79\%), but it also achieves a lower minADE on the original data (-2.88\%). Additionally, for half of the perturbed trajectories minADE improves (Fig.~\ref{fig:per-example}d), and for more than half on original data (Fig.~\ref{fig:per-example}g). On the other hand, MultiPath++ also achieves a lower average performance degradation (+16.03\% vs. +71.64\%), but increased minADE on the original data (+16.42\%). Fig.~\ref{fig:example-predictions}f shows that despite missing road information, there is sufficient information to make significantly better predictions, suggesting that the models were unnecessarily reliant in road information, and learn to better exploit other information in the scene applying this perturbation.

\subsubsection{Late detections}
LSTM achieves a marginally improved performance degradation on perturbed data (+81.24\% vs. +82.95\%) and a marginal performance degradation on the original data (+0.19\%). MotionCNN achieves a significant improvement on perturbed data (+30.60\% vs. +184.89\%), and lower minADE on the original data (-3.6\%). MultiPath++ maintains a similar degradation on perturbed data (87.5\% vs. 85.82\%), but also lower minADE on the original data (\mbox{-4.48\%}). Simulating late detections during training proves beneficial unless past observations are the only additional information the model can leverage. 

\subsubsection{Noisy heading angle}
Introducing this perturbation during training yields a significant improvement for all models, but in some cases at the expense of higher minADE with the original data. Performance degradation on perturbed data for LSTM is lowered from +354.56\% to +16.88\%, and for MotionCNN from +1248.2\% to +0.69\%. However, minADE increases approximately 4.4\% for both models when using original data. On the contrary, MultiPath++ not only achieves a performance improvement on perturbed data (-1.5\% vs. +1444.78\%), but also slightly lower minADE on original data (-0.75\%). For more than half of the trajectories, predicting with the retrained MotionCNN and MultiPath++ using perturbed trajectories yields higher accuracy than using the original trajectories~(Fig. \ref{fig:per-example}, bottom).

\subsection{Are Perturbations the Solution to Achieve Robustness?}
Perturbations are an effective strategy to improve model robustness. Examples from our analysis are shown in Fig.~\ref{fig:example-predictions}. For instance, the removal of road information initially causes off-road predictions (Fig.~\ref{fig:example-predictions}b), although there seems to be sufficient information in historical data to produce better predictions even without any road information (Fig.~\ref{fig:example-predictions}f). Another example is the removal of historical data leading to off-road predictions (Fig.~\ref{fig:example-predictions}c) despite road information being available, which is prevented after simulating the same perturbation during training (Fig.~\ref{fig:example-predictions}d).

Training with perturbed data for the identified perturbations effectively mitigates performance degradation, although it does not address the fact that if the model is presented with previously unseen data, its behavior can be erratic once again. Guaranteeing robustness with perturbations would therefore require identification of all possible complications to simulate the necessary perturbations, which is likely unfeasible. 

Additionally, there are a significant number of cases for which the perturbed data leads to better predictions. For instance, when removing the road with models that were always trained with road (Fig.~\ref{fig:per-example}a). Retraining the models without road not only improves performance degradation but also increases the number of cases with better predictions without road (Fig.~\ref{fig:per-example}d), suggesting the models may be unable to learn the correct relationships in the data.

\section{Conclusions \& Future Work}\label{sec:conclusion}
To enable better trajectory prediction of other road users in real-life experiments, in this work we have compared several models against three types of perturbations we observed when deploying our models in a vehicle, and found that their predictions are severely degraded with minADE increases of up to +110.8\% when road information is unavailable, up to +184.9\% with late detections, and up to +1444.8\% when the last observed heading angle contains high errors.

We show that introducing similar perturbations during model training is effective to mitigate erratic predictions, leading to much lower minADE increases of up to +17.8\% when road information is unavailable, up to +87.5\% with late detections, and up to +16.9\% with noisy heading angles. Despite its effectiveness, this approach requires identification of relevant perturbations before model deployment, and there will likely be new unforeseen circumstances that were not considered and lead to erratic predictions again. Thus, preemptively introducing these perturbations, while effective, does not ensure model robustness. 

Future work will focus on three aspects. Firstly, a comprehensive analysis of  how robustness improvement is affected by the severity of the perturbations and the proportion of perturbed data used during retraining.
Secondly, an evaluation of the in-vehicle robustness enhancement after retraining with simulated perturbations. 
Lastly, further examination of cases where degraded inputs yield improved predictions, which suggest the models do not learn the right relationships between the elements in the scene.

\bibliography{referencesCustom}     
\flushend

\end{document}